%% file: main.tex
\documentclass[10pt,twocolumn,letterpaper]{article}

\usepackage{cvpr}      
\usepackage{microtype} 
\usepackage{xcolor}
\definecolor{myblue}{RGB}{0,114,189} 

\input{preamble}

\definecolor{cvprblue}{rgb}{0.21,0.49,0.74}
\usepackage[pagebackref,breaklinks,colorlinks,citecolor=cvprblue]{hyperref}
\usepackage{multirow}
\usepackage{amsmath}
\usepackage{floatrow}

\def\paperID{134} 
\def\confName{3DV\xspace}
\def\confYear{2026\xspace}

\title{HuPrior3R: Incorporating Human Priors for Better 3D Dynamic
Reconstruction from Monocular Videos}

\author{
Weitao Xiong$^{1,2}$\thanks{Equal contribution.}\quad\quad\quad
Zhiyuan Yuan$^{3}$\footnotemark[1]\quad\quad\quad
Jiahao Lu$^{1}$\\[0.25em]
Chengfeng Zhao$^{1}$\quad\quad\quad
Peng Li$^{1}$\quad\quad\quad
Yuan Liu$^{1}$\thanks{Corresponding author.}\\[1em]
$^{1}$HKUST\quad\quad\quad\quad
$^{2}$XMU \quad\quad\quad\quad
$^{3}$SYSU
}

\begin{document}
\twocolumn[{%
\renewcommand\twocolumn[1][]{#1}%
\maketitle
\begin{center}
    \centering
    \captionsetup{type=figure}
    \includegraphics[width=0.95\textwidth]{sec/figure/teaser.pdf}
    \caption{\textbf{HuPrior3R for Dynamic Scene Reconstruction.} Our task is to reconstruct accurate 3D point clouds from monocular videos containing humans. By incorporating SMPL human priors, HuPrior3R achieves superior reconstruction quality for both dynamic human motions and static scene elements compared to existing methods. The highlighted regions demonstrate our method's ability to preserve anatomically consistent human geometry and fine-grained details while maintaining overall scene coherence.}

    \label{fig:teaser}
\end{center}%
}]
\begingroup
    \renewcommand\thefootnote{\fnsymbol{footnote}}
    \footnotetext[1]{Equal contribution.}
    \footnotetext[2]{Corresponding author.}
\endgroup
\input{sec/0_abstract} 
\input{sec/1_intro}
\input{sec/2_related_work}
\input{sec/3_method}

\input{sec/4_experiment}

\input{sec/5_conclusion}
{
    \small
    \bibliographystyle{ieeenat_fullname}
    \bibliography{main}
}
\input{sec/X_suppl}

\end{document}

%% file: preamble.tex
\usepackage[dvipsnames]{xcolor}


%% file: sec/0_abstract.tex
\begin{abstract}

Monocular dynamic video reconstruction faces significant challenges in dynamic human scenes due to geometric inconsistencies and resolution degradation issues. Existing methods lack 3D human structural understanding, producing geometrically inconsistent results with distorted limb proportions and unnatural human-object fusion, while memory-constrained downsampling causes human boundary drift toward background geometry. To address these limitations, we propose to incorporate hybrid geometric priors that combine SMPL human body models with monocular depth estimation. Our approach leverages structured human priors to maintain surface consistency while capturing fine-grained geometric details in human regions. We introduce HuPrior3R, featuring a hierarchical pipeline with refinement components that processes full-resolution images for overall scene geometry, then applies strategic cropping and cross-attention fusion for human-specific detail enhancement. The method integrates SMPL priors through a Feature Fusion Module to ensure geometrically plausible reconstruction while preserving fine-grained human boundaries. Extensive experiments on TUM Dynamics and GTA-IM datasets demonstrate superior performance in dynamic human reconstruction.

\end{abstract}

%% file: sec/1_intro.tex
\section{Introduction}
\label{sec:intro}
Recovering both camera motion and geometry from monocular dynamic video is still a challenging task in computer vision, with applications ranging from robotics to augmented reality.
Traditional methods utilize Structure-from-Motion (SFM), which involve iterative optimization techniques, such as Bundle Adjustment (BA), to recover the static scene structure.
However, these methods are limited by their reliance on feature correspondences, which are often sparse.
%
%
Recent methods~\cite{wang2024dust3r} address this problem in a data-driven manner,
where DUSt3R~\cite{wang2024dust3r} bypasses the need for explicit feature matching by directly learning scene geometry and motion from image pair input, enabling faster, more accurate, and more robust handling of the dense 3D geometry.
However, they are specifically designed for static scenes, which cannot be directly generalized to monocular dynamic video input.
Prior works have proposed several methods for handling dynamic scenes.
MonST3R~\cite{zhang2025monst3rsimpleapproachestimating} fine-tunes the DUSt3R model on dynamic data and introduces a new optimization method specifically designed for dynamic scenes.
Align3R~\cite{lu2024align3ralignedmonoculardepth} enhances DUSt3R~\cite{wang2024dust3r} by incorporating a depth prior, improving its performance on monocular dynamic video input.
Meanwhile, 
Easi3R~\cite{chen2025easi3restimatingdisentangledmotion} extends DUSt3R to dynamic environments by disentangling attention maps to separate dynamic and static motion.
VGGT~\cite{wang2025vggtvisualgeometrygrounded} addresses this challenge by jointly training a large transformer on both static and dynamic data, allowing for better generalization across diverse scenarios.
While these methods demonstrate strong performance on general dynamic scenes, they still have two issues in the domain of precise human depth estimation.
%
\textit{1. Anatomical Artifacts:} Due to the lack of adequate 3D human structural understanding, existing methods often produce geometrically inconsistent results—such as distorted limb proportions, disconnected body parts, and unnatural human-object fusion. 
\textit{2. Resolution and Boundary Degradation:} Humans typically occupy small pixel regions in scene images, and memory-constrained image downsampling (from $1080 \times 1920$ to $288 \times 512$) exacerbates this issue by causing ViT tokenization tends to blend foreground human and background information within the same spatial tokens. Consequently, reconstructed human boundaries drift toward background geometry, losing fine-grained human details. 
%

To address the aforementioned challenges, we propose a novel method, \textbf{HuPrior3R}.
To mitigate the Anatomical Artifacts issue, we introduce a hybrid geometric prior that combines the Skinned Multi-Person Linear (SMPL) model with monocular depth estimation. Specifically, the SMPL model provides a consistent human surface prior, while monocular depth estimation contributes fine-grained geometric details. By aligning the SMPL geometry with the monocular depth map, we obtain detailed depth predictions that maintain strong surface consistency. Furthermore, to tackle the Resolution and Boundary Degradation issue, we design a pipeline with a refinement component. The network initially processes the full-resolution image to capture the overall scene geometry. When humans appear small in the image, our refinement module applies strategic image cropping to effectively super-resolve human regions. A dedicated branch then employs cross-attention between the enlarged human region and the original image context, enabling precise reconstruction of human-like forms and ultimately producing more detailed and accurate human reconstruction results.
\begin{figure*}[t]
\vspace{-2em}
  \centering
    \includegraphics[width=1\linewidth]{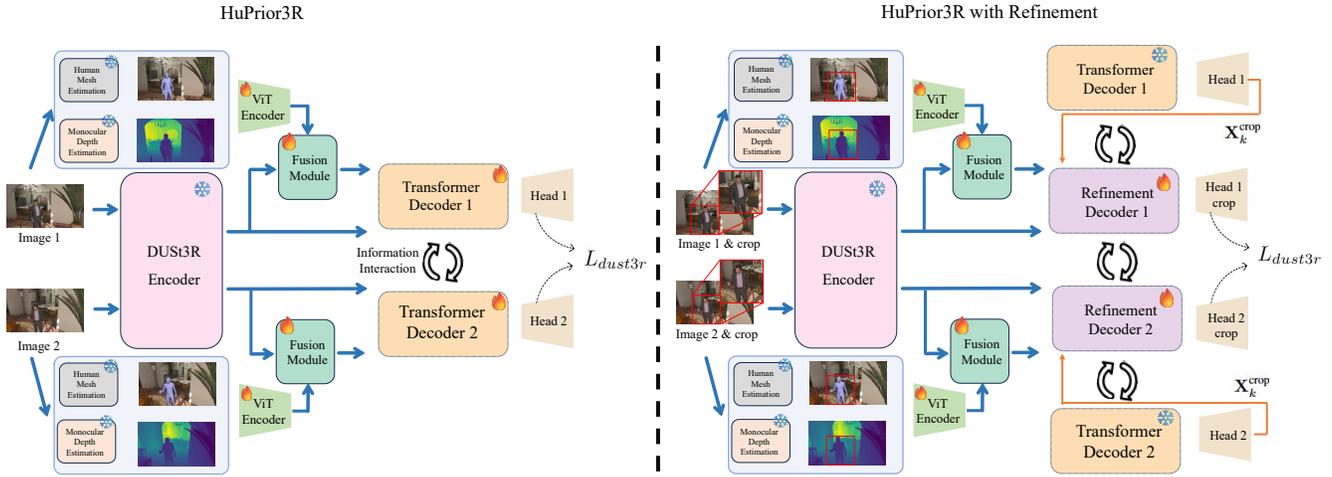}
    \caption{\textbf{The overview of our framework.} The left shows the base HuPrior3R pipeline that processes image pairs through DUSt3R encoders, followed by a Feature Fusion Module that integrates image features, monocular depth features, and SMPL depth features via cross-attention mechanism and gated operations. The fused features are then processed through transformer decoders to produce initial point map reconstruction. The right illustrates HuPrior3R with Refinement, which is activated when humans occupy small pixel regions in the input images. The refinement stage crops human regions from the input images, upsamples the point clouds, and processes them through refinement decoders that incorporate cross-module attention between local crop features and preserved scene context. This hierarchical design enables high-fidelity human reconstruction while maintaining geometric consistency between human regions and the surrounding scene.}
\end{figure*}

Our main contributions are summarized as follows:
\begin{itemize}
\item We propose a novel model, \textbf{HuPrior3R}, capable of accurate dynamic human reconstruction.
\item To address the Anatomical Artifacts issue, we incorporate a hybrid geometric prior that combines the SMPL model with monocular depth, enabling detailed depth prediction while maintaining strong surface consistency.
\item To mitigate the Resolution and Boundary Degradation issue, we design a hierarchical pipeline with a refinement component: the network initially processes full-resolution images to capture overall scene geometry, then applies a dedicated cropping mechanism and cross-attention fusion to refine human-specific details when humans appear small in the scene.
\item Extensive experiments demonstrate the effectiveness of our approach, with results on the \textbf{TUM dynamics} and \textbf{GTA-IM} datasets validating its strong performance in dynamic human reconstruction.
\end{itemize}

%% file: sec/2_related_work.tex
\section{Related Work}
\label{sec:related_work}

\paragraph{Feed-Forward 3D Reconstruction}
Recent advancements in feed-forward 3D reconstruction~\cite{wang2024dust3r, wang20243dreconstructionspatialmemory, tang2024mvdust3rsinglestagescenereconstruction, Elflein_2025_CVPR,cao2025reconstructing, cabon2025must3rmultiviewnetworkstereo, jang2025pow3rempoweringunconstrained3d, liu2025regist3rincrementalregistrationstereo, maggio2025vggtslamdensergbslam, li2025mono3rexploitingmonocularcues} have focused on directly regressing dense scene geometry from image inputs without relying on iterative optimization. DUSt3R~\cite{wang2024dust3r} introduces a dense and unconstrained stereo reconstruction framework that predicts local pointmaps from image pairs without known camera poses or calibration, achieving global consistency via lightweight alignment. Recent extensions such as ~\cite{wang20243dreconstructionspatialmemory,tang2024mvdust3rsinglestagescenereconstruction, cabon2025must3rmultiviewnetworkstereo, Elflein_2025_CVPR} eliminate DUSt3R’s dependence on global optimization. While these approaches offer efficient and accurate reconstruction of static environments, they are not designed for dynamic scenes, particularly those containing human subjects.
\vspace{-0.2em}
\paragraph{Feed-Forward 4D Reconstruction}
Recent feed-forward 4D reconstruction methods aim to recover time-varying scene geometry from monocular dynamic video~\cite{cao2025reconstructing}. MonST3R~\cite{zhang2025monst3rsimpleapproachestimating} extends DUSt3R by introducing temporal consistency losses and re-optimization strategies to better capture dynamic motion. CUT3R~\cite{wang2025continuous3dperceptionmodel} addresses dynamic reconstruction by using a stateful recurrent transformer to accumulate 3D information across frames into a persistent spatial memory, enabling online and globally aligned 4D reconstruction. Easi3R~\cite{chen2025easi3restimatingdisentangledmotion} improves generalization to deformable objects by disentangling appearance and motion via attention mechanisms. VGGT~\cite{wang2025vggtvisualgeometrygrounded, deng2025vggtlongchunkitloop, wang2025pi3scalablepermutationequivariantvisual} and Rig3R~\cite{li2025rig3rrigawareconditioninglearned} trains large transformer models jointly on static and dynamic datasets to enhance overall robustness. Uni4D~\cite{yao2025uni4dunifyingvisualfoundation} introduces a multi-stage optimization framework that harnesses multiple pretrained models to advance dynamic 3D modeling. Geo4D~\cite{jiang2025geo4dleveragingvideogenerators} repurposes video diffusion models to predict multi-modal geometric outputs—point, depth, and ray maps—and fuses them at test time for accurate long-range dynamic scene reconstruction. GeometryCrafter~\cite{xu2025geometrycrafterconsistentgeometryestimation}, on the other hand, uses a point-map VAE and diffusion-based sequence modeling to estimate high-fidelity, temporally coherent point map sequences from real-world videos. Among these, Align3R~\cite{lu2024align3ralignedmonoculardepth} offers a simple yet effective strategy by injecting depth prior into DUSt3R’s pointmap framework, improving details and temporal consistency in dynamic environments. However, these methods treat human regions as unstructured geometry and overlook the blended information between the foreground and background, often leading to anatomical artifacts and boundary degradation. MegaSAM~\cite{li2024megasamaccuratefastrobust} and VIPE~\cite{huang2025vipe} combine the advantages of feed forward model and post-alignment techniques, demonstrating impressive results. Others method like ~\cite{deng2025vggtlongchunkitloop, wang2025vggtvisualgeometrygrounded, feng2025st4rtracksimultaneous4dreconstruction, zhang2025tapip3dtrackingpointpersistent, cho2025seuratmovingpointsdepth, chen2025trackbundleadjustmentdynamic, zhang2025pomatomarryingpointmapmatching, han2025d2ust3renhancing3dreconstruction, liang2025zeroshotmonocularsceneflow, xiao2025spatialtrackerv23dpointtracking} utilize 2D/3D tracks or motion prediction~\cite{lin2025moviesmotionaware4ddynamic} to improve the reconstruction quality.
In contrast, our method builds upon Align3R, incorporating a hybrid geometric prior and proposing a refinement strategy to achieve anatomically coherent and address the boundary degradation.
\vspace{-1.2em}
\paragraph{Monocular Depth Estimation}
Monocular depth estimation methods such as MiDaS~\cite{ranftl2021visiontransformersdenseprediction}, DepthAnything~\cite{yang2024depthanythingunleashingpower}, and DepthPro~\cite{bochkovskii2025depthprosharpmonocular} have demonstrated strong performance in producing accurate single-image depth maps through large-scale training on both labeled and unlabeled data. MiDaS~\cite{ranftl2021visiontransformersdenseprediction} mixes diverse datasets to achieve robust zero-shot generalization across domains, while DepthAnything~\cite{yang2024depthanythingunleashingpower} scales up to millions of unlabeled images for improved robustness. DepthPro~\cite{bochkovskii2025depthprosharpmonocular} further incorporates metric-scale supervision and sharp-detail modeling for precise depth predictions. Additional works~\cite{bhat2023zoedepthzeroshottransfercombining,ranftl2021visiontransformersdenseprediction,Farooq_Bhat_2021,gabdullin2025depthartmonoculardepthestimation,Gasperini_2023} explore directions such as improved zero-shot transfer, dense transformer architectures, adaptive binning, autoregressive refinement, and enhanced robustness under challenging conditions. While monocular depth estimation generally provides detailed and high-quality geometric cues, it is inherently frame-independent and thus prone to temporal inconsistency and flickering when applied to video sequences—issues that are especially pronounced in dynamic scenes involving humans. Moreover, monocular depth methods often produce degraded predictions on human surfaces, leading to anatomical artifacts such as distorted limbs, inconsistent body contours, and missing fine-scale geometry. These challenges motivate the need for approaches that preserve detailed geometry, ensure surface consistency, and maintain temporal stability in dynamic human reconstruction. 
\vspace{-0.6em}
\paragraph{Human Mesh Estimation and Recovery}
Recent human mesh recovery approaches~\cite{kanazawa2018endtoendrecoveryhumanshape,goel2023humans4dreconstructingtracking,kolotouros2019learningreconstruct3dhuman,dwivedi2024tokenhmradvancinghumanmesh,yuan2022glamrglobalocclusionawarehuman} have made significant progress by leveraging strong backbones and large-scale 3D datasets. Most of these methods, including HMR~\cite{kanazawa2018endtoendrecoveryhumanshape}, SPIN~\cite{kolotouros2019learningreconstruct3dhuman}, and HMR2.0~\cite{goel2023humans4dreconstructingtracking}, adopt a weak-perspective camera model to simplify training, which however leads to scale ambiguity in real-world scenes. More recent works such as TokenHMR~\cite{dwivedi2024tokenhmradvancinghumanmesh} and GLAMR~\cite{yuan2022glamrglobalocclusionawarehuman} highlight the importance of accurate camera modeling and spatiotemporal consistency, yet still suffer from misalignment when projecting 3D poses into image space.
CameraHMR~\cite{patel2024camerahmraligningpeopleperspective} addresses this issue by learning full-perspective camera parameters during training and inference, significantly improving scale stability and geometric alignment. Given that our method relies on accurate depth priors and consistent human geometry across frames, we adopt CameraHMR to obtain reliable SMPL estimates under dynamic camera motions.

%% file: sec/3_method.tex
\section{Method}
\label{sec:method}



\subsection{Overview}
Given a video consisting of $N$ frames $\{\mathbf{I}_k \in \mathbb{R}^{H \times W \times 3}\}_{k=1}^N$, our target is to estimate the corresponding depth maps $\{\mathbf{D}_k \in \mathbb{R}^{H \times W}\}_{k=1}^N$ and camera poses $\{\pi_k \in \mathbb{SE}(3)\}_{k=1}^N$. HuPrior3R achieves this through a hierarchical pipeline with a refinement:

\begin{itemize}
\item\textbf{Global Scene Processing.} We estimate coarse depth maps for all frames and generate global point clouds by fusing monocular cues with parametric human priors via a feature fusion module.
\item\textbf{Human-Centric Refinement.} When humans appear small in the scene, we crop both the input images and the initial point clouds to the localized human regions. A dedicated network branch then performs cross-attention between the enlarged human region and the original image context. This design facilitates more accurate point cloud alignment and subsequent refinement, ensuring proper human-like form reconstruction.
\end{itemize}

\subsection{Global Scene Processing}
\subsubsection{Aligning Depth with SMPL Priors}
To ensure consistent depth representation across different estimation sources, we align the monocular depth maps $\{\hat{\mathbf{D}}_k^{\text{Mono}}\}_{k=1}^N$ with the SMPL depth maps $\{\hat{\mathbf{D}}_k^{\text{SMPL}}\}_{k=1}^N$ via a robust linear fitting scheme based on RANSAC. For each frame $k$, a binary human mask $\mathbf{M}_k \in \{0,1\}^{H \times W}$ is obtained from the 2D SMPL pose to define valid human regions. Given a monocular depth map $\hat{\mathbf{D}}_k^{\text{Mono}}$, only the depth values inside the valid mask are extracted to exclude background pixels. Let $\mathbf{d}_{\text{mono}} \in \mathbb{R}^{n}$ and $\mathbf{d}_{\text{smpl}} \in \mathbb{R}^{n}$ denote the flattened depth vectors from $\hat{\mathbf{D}}_k^{\text{Mono}}$ and $\hat{\mathbf{D}}_k^{\text{SMPL}}$, respectively, at the $n$ valid pixel locations. RANSAC is then applied to estimate the scale factor $s$ and offset $b$ of the optimal linear transformation:
\begin{equation}
    \mathbf{d}_{\text{smpl}}(i) \approx s \cdot \mathbf{d}_{\text{mono}}(i) + b, \quad i = 1, \ldots, n,
\end{equation}
where $\mathbf{d}_{\text{smpl}}(i)$ and $\mathbf{d}_{\text{mono}}(i)$ are the depth values at the $i$-th valid pixel.
The RANSAC algorithm proceeds by iteratively sampling random subsets of two points from $\mathbf{d}_{\text{mono}}$ and $\mathbf{d}_{\text{smpl}}$, fitting a linear model, and evaluating the number of inliers within a predefined residual threshold $\tau$ (set to 0.001 in our implementation). The model with the highest number of inliers after a fixed number of iterations (e.g., 1000) is selected, yielding the best scale $s^*$ and shift $b^*$.

Finally, the aligned monocular depth map $\hat{\mathbf{D}}_k^{\text{Aligned}}$ is obtained by applying the transformation to the entire $\hat{\mathbf{D}}_k^{\text{Mono}}$ as follows:

\begin{equation}
\hat{\mathbf{D}}_k^{\text{Aligned}}(x, y) = s^* \cdot \hat{\mathbf{D}}_k^{\text{Mono}}(x, y) + b^*,
\end{equation}
for all pixel coordinates $(x, y)$. Additionally, we leverage the focal length predicted by the Human Mesh Estimation model for the SMPL representation as a reference in point map reconstruction. The aligned depth map, $\hat{\mathbf{D}}_k^{\text{Aligned}}$, and the SMPL depth map, $\hat{\mathbf{D}}_k^{\text{SMPL}}$, are subsequently unprojected into point maps, $\hat{\mathbf{X}}_k^{\text{Aligned}}$ and $\hat{\mathbf{X}}_k^{\text{SMPL}}$, using the SMPL focal length.

\subsubsection{Feature Fusion Module}
After obtaining the point maps from the aligned monocular depth map $\hat{\mathbf{D}}_k^{\text{Aligned}}$ and the SMPL depth map $\hat{\mathbf{D}}_k^{\text{SMPL}}$, we aim to fuse their geometric and contextual information to enhance the representation of the human body in 3D space. To achieve this, we employ a cross-attention mechanism to integrate features extracted from both point maps, followed by a gated operation to adaptively control the contribution of the fused features. This Feature Fusion Module is applied before the decoder in both Stage 1 and Stage 2 of our pipeline.

\textbf{Point map ViT.} We first extract point map features $\hat{\mathbf{F}}_{\text{mono}}$ and $\hat{\mathbf{F}}_{\text{smpl}}$ using trainable lightweight ViT-based encoders. To process the point maps, we apply a patch embedding method to divide the input point maps into patches and embed them into feature tokens, formulated as:
\begin{equation}
\hat{\mathbf{F}}_i = \text{PatchEmbed}(\hat{\mathbf{X}}_i),
\end{equation}
where $\hat{\mathbf{X}}_i$ represents the input point map, and $\hat{\mathbf{F}}_i$ is the corresponding patchified feature representation with shape dependent on the patch size and embedding dimension.

\textbf{Cross Attention Module.} To maintain DUSt3R’s robust feature extraction capabilities, the image encoder is frozen. image features $\hat{\mathbf{F}}_{\text{img}}$ are extracted from the input views using this frozen ViT encoder. A cross-attention module is then applied to integrate above features. Specifically, the query is formed by concatenating $\hat{\mathbf{F}}_{\text{img}}$ and $\hat{\mathbf{F}}_{\text{mono}}$ followed by a linear projection, while $\hat{\mathbf{F}}_{\text{smpl}}$ serves as both the key and value after a separate projection. This mechanism allows the module to attend to relevant geometric cues in the SMPL depth features, producing an output feature set $\hat{\mathbf{F}}_{\text{cross}}$ that captures interdependencies among the image, monocular depth, and SMPL depth features. The cross-attention output is computed as:
\begin{equation}
\hat{\mathbf{F}}_{\text{cross}} = \textit{MultiAtten}(\textit{\textbf{Q}}(\hat{\mathbf{F}}_{\text{img}}, \hat{\mathbf{F}}_{\text{mono}}), \textit{\textbf{K}}(\hat{\mathbf{F}}_{\text{smpl}}), \textit{\textbf{V}}(\hat{\mathbf{F}}_{\text{smpl}})),
\end{equation}
where $\textit{\textbf{Q}uery}(\cdot)$, $\textit{\textbf{K}ey}(\cdot)$, and $\textit{\textbf{V}alue}(\cdot)$ denote the respective projections, and $\textit{Multi-Atten}(\cdot)$ represents the multi-head attention mechanism.

\textbf{Gate Mechanism.} To adaptively integrate $\hat{\mathbf{F}}_{\text{cross}}$ with the original monocular features $\hat{\mathbf{F}}_{\text{mono}}$, we use a gate mechanism to compute a gate $\mathbf{G}$, which determines the contribution of $\hat{\mathbf{F}}_{\text{cross}}$. The gate is derived from the concatenation of $\hat{\mathbf{F}}_{\text{cross}}$, $\hat{\mathbf{F}}_{\text{mono}}$, and $\hat{\mathbf{F}}_{\text{smpl}}$ through a series of linear transformations and activations. The final fused feature is obtained as:
\vspace{-0.3em}
\begin{equation}
\hat{\mathbf{F}}_{\text{fused}} = \hat{\mathbf{F}}_{\text{mono}} + \mathbf{G} \odot \hat{\mathbf{F}}_{\text{cross}},
\end{equation}
where $\odot$ denotes element-wise multiplication. This gated integration ensures that the cross-attention features selectively enhance the original monocular features, preserving critical geometric information while incorporating complementary cues from the SMPL depth map.

\subsubsection{Decoder Module}
We process the full scene images $\{\mathbf{I}_k\}_{k=1}^N$ to establish global context. The fused features $\hat{\mathbf{F}}_{\text{fused}}$ from the Feature Fusion Module are processed through self-attention to generate multi-level features $\hat{\mathbf{F}}_{\text{fused}}^{(1)}, \hat{\mathbf{F}}_{\text{fused}}^{(2)}, \ldots, \hat{\mathbf{F}}_{\text{fused}}^{(s)}$, which are injected into the DUSt3R decoder via zero convolution:
\begin{equation}
\hat{\mathbf{E}}_{\text{scene}}^{(l)} = \text{ZeroConv}(\hat{\mathbf{F}}_{\text{fused}}^{(l)}) + \mathbf{E}_{\text{base}}^{(l)}, \quad l = 1, 2, \ldots, s
\end{equation}
where $\mathbf{E}_{\text{base}}^{(l)}$ represents the original DUSt3R decoder features. This produces global low-resolution point maps $\{\mathbf{X}_k\}_{k=1}^N$ while preserving the enhanced intermediate decoder features  $\{\hat{\mathbf{E}}_{\text{scene}}^{(l)}\}_{l=1}^s$ for potential refinement when humans appear small in the scene.

\subsection{Human-Centric Refinement}

When humans occupy small pixel regions in input images, we activate a refinement module to address resolution limitations and boundary inaccuracies from the global processing. This refinement enables precise depth and geometry estimation for human-centric regions by leveraging both global scene context and local geometric details.

\subsubsection{Input Cropping and Preparation}  
We begin by extracting bounding boxes from the SMPL depth maps to crop human regions from the input images $\{\mathbf{I}_k^{\text{crop}}\}_{k=1}^N$, monocular depth maps, and global point clouds $\{\mathbf{X}_k^{\text{crop}}\}_{k=1}^N$. $\mathbf{X}_k^{\text{crop}}$ are then upsampled via bicubic interpolation to match the resolution of the cropped image regions, providing dense spatial $\hat{\mathbf{F}}_{\mathrm{point}}$. In parallel, we preserve multi-scale decoder features $\{\hat{\mathbf{F}}_{\text{scene}}^{(l)}\}_{l=2}^s$ from the global processing, which encode valuable scene context for cross-module integration.

The cropped images and monocular depth maps are processed through the same encoder and Feature Fusion Module as in global processing, generating multi-level crop-specific features $\{\hat{\mathbf{F}}_{\text{crop}}^{(l)}\}_{l=1}^{s}$ enriched with high-frequency local geometry.

\subsubsection{Cross-Module Decoder Integration}  
To effectively combine global and local information, the refinement decoder incorporates cross-attention at each decoding layer. The integration process follows a two-step approach:

\textbf{Initial Layer Integration:} At the first decoder layer ($l=1$), we integrate spatial guidance from upsampled point clouds:
\begin{equation}
\hat{\mathbf{E}}_{\mathrm{crop}}^{(1)} = \mathrm{ZeroConv}(\hat{\mathbf{F}}_{\mathrm{crop}}^{(1)}) + \mathrm{ZeroConv}(\hat{\mathbf{F}}_{\mathrm{point}}) + \hat{\mathbf{E}}_{\mathrm{crop\,base}}^{(1)}
\end{equation}
\textbf{Cross-Module Attention:} For subsequent layers ($l = 2, \ldots, s$), we apply cross-attention between local crop features and preserved global scene context:
\begin{equation}
\hat{\mathbf{F}}_{\mathrm{crop}}^{\prime(l)}= \textit{CrossAttn}(\hat{\mathbf{F}}_{\mathrm{crop}}^{(l)}, \hat{\mathbf{F}}_{\text{scene}}^{(l)})
\end{equation}
\vspace{-0.5em}
\begin{equation}
\hat{\mathbf{E}}_{\mathrm{crop}}^{(l)} = \mathrm{ZeroConv}(\hat{\mathbf{F}}_{\mathrm{crop}}^{\prime(l)}) + \hat{\mathbf{E}}_{\mathrm{crop\,base}}^{(l)}
\end{equation}

where $\hat{\mathbf{F}}_{\mathrm{crop}}^{(l)}$ provides human features from the cropped pointmap region, $\hat{\mathbf{F}}_{\text{scene}}^{(l)}$ introduces global context preserved from scene processing, $\hat{\mathbf{F}}_{\mathrm{point}}$ represents spatial guidance from upsampled point clouds, and $\hat{\mathbf{E}}_{\mathrm{crop\,base}}^{(l)}$ denotes the intermediate hidden representations from the refinement decoder itself. This formulation ensures that the decoder remains aware of both fine-scale human shapes and the global scene layout.

\textbf{Output Integration.}  
The refinement decoder produces high-resolution point maps for each cropped region. Thanks to the use of spatial guidance and consistent camera intrinsics, the output is inherently aligned with the global coordinate system. We resize the refined point maps back to the original resolution and integrate them into the global point clouds through spatial alignment in the image domain. This design enables high-fidelity human reconstruction while maintaining geometric consistency between human regions and the surrounding scene.

\subsection{Fine-tuning on Human Dynamic Videos}
Both global processing and human-centric refinement use the same fine-tuning loss following DUSt3R: 
\begin{equation}
L_{dust3r}=\mathbf{C}_v^e\left\|\frac{1}{z}\mathbf{X}_v^e-\frac{1}{\overline{z}}\overline{\mathbf{X}}_v^e\right\|_2-\alpha \log \mathbf{C}_v^e
\end{equation}
where $v\in\{n,m\}$ denotes the view index, $\mathbf{X}$ and $\overline{\mathbf{X}}$ are the predicted and ground-truth point maps, $\mathbf{C}_v^e$ is the correspondence confidence map, $z$ and $\overline{z}$ are scaling factors used to normalize the predicted and ground truth point maps, and $\alpha$ is the same hyperparameter as in DUSt3R. Notably, global processing and refinement use the same $z$ and $\overline{z}$ for consistency.

%% file: sec/4_experiment.tex
\section{Experiment}
\subsection{Experimential Setting} 
\paragraph{Training datasets.} For fine-tuning, we use two synthetic datasets and one real-world dataset. The synthetic datasets are Bedlam \cite{Black_CVPR_2023} and GTA-IM \cite{caoHMP2020}, while the real dataset is BEHAVE \cite{bhatnagar22behave}.  All datasets are downsampled to 5 FPS.
\vspace{-0.5em}
\paragraph{Training details.} Our HuPrior3R is initialized with the weight of Align3R \cite{lu2024align3ralignedmonoculardepth} and trained using AdamW \cite{Loshchilov2017DecoupledWD} with a learning rate of $5 \times 10^{-5}$ and a batch size of 5 on 7 NVIDIA H800 GPUs. Training our model on $5 \times 10^{4}$ pairs for 50 epochs in each of main pipeline and refinement separately takes approximately one week in total.
\subsection{Video Depth Estimation}
\paragraph{Evaluating dataset.}For evaluation, we use five datasets, comprising three real-world datasets—Bonn~\cite{10.1109/IROS40897.2019.8967590}, TUM Dynamics~\cite{sturm12iros}, and BEHAVE~\cite{bhatnagar22behave}, and two synthetic datasets, GTA-IM and Bedlam. For GTA-IM and Bedlam, we ensure that the evaluation scenes and human motions are entirely distinct from those used in the training set. For BEHAVE, we use the designated evaluation subset of the dataset. For Bonn \cite{10.1109/IROS40897.2019.8967590} and TUM Dynamics \cite{sturm12iros}, we select three scenes featuring human subjects.
\paragraph{Evaluation metrics.} For quantitative evaluation, we mainly report two widely used depth metrics: absolute relative error (Abs Rel) and the percentage of inlier points (with a threshold value of $\delta < 1.25$). Furthermore, we follow the evaluation protocol established by MonST3R \cite{zhang2025monst3rsimpleapproachestimating} to ensure a fair comparison with existing methods.
\paragraph{Baselines.} We compare our HuPrior3R with 1 diffusion-based 4D feed forward reconstruction method: \textbf{Geo4D}  \cite{jiang2025geo4dleveragingvideogenerators}, and three non-diffusion-based methods: \textbf{MonST3R} \cite{zhang2025monst3rsimpleapproachestimating}, \textbf{Align3R} \cite{lu2024align3ralignedmonoculardepth}, and \textbf{VGGT} \cite{wang2025vggtvisualgeometrygrounded}.
\begin{table*}[!t]
\vspace{-2em}
  \begin{center}
    \footnotesize
    \setlength\tabcolsep{1.5pt}
    \caption{\textbf{Video depth estimation results.} We evaluate our model on both real-world datasets: Bonn and TUM dynamics, Behave, and synthetic datasets: GTA-IM, and Bedlam. For Bonn and TUM dynamics we just pick scenes with human. \textbf{Best} and \underline{second best} results are highlighted.}
    \label{table:depth}
    \scalebox{0.9}{
    \begin{tabular}{ll|cc|cc|cc|cc|cc}
      \toprule
       \multirow{3}{*}{Category} &\multirow{3}{*}{Method} & \multicolumn{4}{c|}{Synthetic}& \multicolumn{6}{c}{Real-World}\\
       \cline{3-12}
       &&\multicolumn{2}{c|}{GTA-IM} &   \multicolumn{2}{c|}{Bedlam} &  \multicolumn{2}{c|}{BEHAVE} &\multicolumn{2}{c|}{Bonn 3 scenes} &  \multicolumn{2}{c}{TUM dynamics}  \\
       & & Abs Rel $\downarrow$& $\delta<1.25 \uparrow$   & Abs Rel $\downarrow$& $\delta<1.25 \uparrow$& Abs Rel $\downarrow$& $\delta<1.25 \uparrow$& Abs Rel $\downarrow$& $\delta<1.25 \uparrow$& Abs Rel $\downarrow$& $\delta<1.25 \uparrow$ \\
       \midrule
       \multirow{1}{*}{Diffusion based}
       & Geo4D \cite{jiang2025geo4dleveragingvideogenerators} 
            & 0.218 & 0.717 & \textbf{0.058} & \textbf{0.943} & 0.194 & 0.752 & 0.087 & \textbf{0.966} & 0.175 & 0.768 \\
       \midrule
        \multirow{5}{*}{Non-Diffusion}
       & MonST3R~\cite{zhang2025monst3rsimpleapproachestimating}                 & 0.265 & 0.626 & 0.115 & 0.884 &  0.082 & 0.903 & 0.094 & 0.915 & 0.134 & 0.843 \\
       & Align3R~\cite{lu2024align3ralignedmonoculardepth}                 & 0.138 & 0.775 & \underline{0.086} &  \underline{0.935} & \underline{0.049} & \underline{0.966} & \underline{0.086} & 0.938 & 0.104 & 0.890 \\  
       & VGGT~\cite{wang2025vggtvisualgeometrygrounded}
         & \textbf{0.104} & \underline{0.807} & 0.200 & 0.646 & 0.699 & 0.904 & 0.267 & 0.943 & 0.218 & 0.877 \\
       & Ours
         & \underline{0.112} & \textbf{0.869} & 0.107 & 0.855 & \textbf{0.033} & \textbf{0.992} & \textbf{0.077} & \underline{0.959} & \textbf{0.102} & \textbf{0.907} \\
      \bottomrule
    \end{tabular}}
     \vspace{-1.2em}
  \end{center}
\end{table*}
\paragraph{Results.} Table~\ref{table:depth} presents the quantitative results for video depth estimation. Our HuPrior3R demonstrates superior performance on real-world datasets, achieving the best results on BEHAVE (Abs Rel: 0.033, $\delta<1.25$: 0.992), Bonn (Abs Rel: 0.077), and TUM Dynamics (Abs Rel: 0.102, $\delta<1.25$: 0.907). On synthetic datasets, our method shows competitive performance on GTA-IM (second-best Abs Rel: 0.112, best $\delta<1.25$: 0.869), while Geo4D performs better on Bedlam due to its diffusion-based nature. Overall, our approach effectively handles human-centric scenes with complex interactions.

\begin{table*}[!t]
  \begin{center}
    \footnotesize
    \setlength\tabcolsep{2.5pt}
    \caption{\textbf{Camera pose estimation results.} We evaluate our model on two real-world datasets: TUM dynamics and Bonn, as well as the synthetic dataset: GTA-IM. \textbf{Best} and
\underline{second best} results are highlighted.
}
    \label{pose_estimation}

    \begin{tabular}{ll|ccc|ccc|ccc}
      \toprule
       \multirow{2}{*}{Category} &\multirow{2}{*}{Method} &  \multicolumn{3}{c|}{GTA-IM} &\multicolumn{3}{c|}{Bonn 3 scenes}&  \multicolumn{3}{c}{TUM dynamics}  \\

       & & ATE $\downarrow$& RTE$\downarrow$   & RRE$\downarrow$& ATE($10^{-2}$) $\downarrow$&RTE($10^{-2}$)$\downarrow$   & RRE$\downarrow$ & ATE($10^{-2}$) $\downarrow$&RTE($10^{-2}$)$\downarrow$   & RRE$\downarrow$\\
    \midrule
    \multirow{1}{*}{Diffusion based}
    & Geo4D \cite{jiang2025geo4dleveragingvideogenerators} 
            & 0.107 & 0.222 & 5.66 & \textbf{0.367} & 0.321 & \textbf{0.282} & 0.461 & \underline{0.269} & 0.480\\
       \midrule
    \multirow{4}{*}{Non-Diffusion}
     &MonST3R~\cite{zhang2025monst3rsimpleapproachestimating}& 0.039 &  0.160 & \underline{3.23} & 0.499 & \textbf{0.276} & 0.521 & 0.407 & 0.271 & 0.506\\
    &Align3R~\cite{lu2024align3ralignedmonoculardepth}& \underline{0.036} & \underline{0.120} & 6.28 & 0.519 & 0.298 & 0.510 & \textbf{0.295} & \textbf{0.240} & \underline{0.432} \\
    &VGGT~\cite{wang2025vggtvisualgeometrygrounded}& 0.292 & 0.292 & 8.69 & \underline{0.460} & 0.544 & 0.535 & 0.337 & 0.294 & 0.591 \\
    &Ours& \textbf{0.015} & \textbf{0.010} & \textbf{0.227} & 0.473 & \underline{0.285} & \underline{0.503} & \underline{0.335} & 0.290 & \textbf{0.430}\\
      \bottomrule
    \end{tabular}
     \vspace{-1.2em}
  \end{center}
\end{table*}

\begin{figure*}[t]
    \centering
    \includegraphics[width=0.94\linewidth]{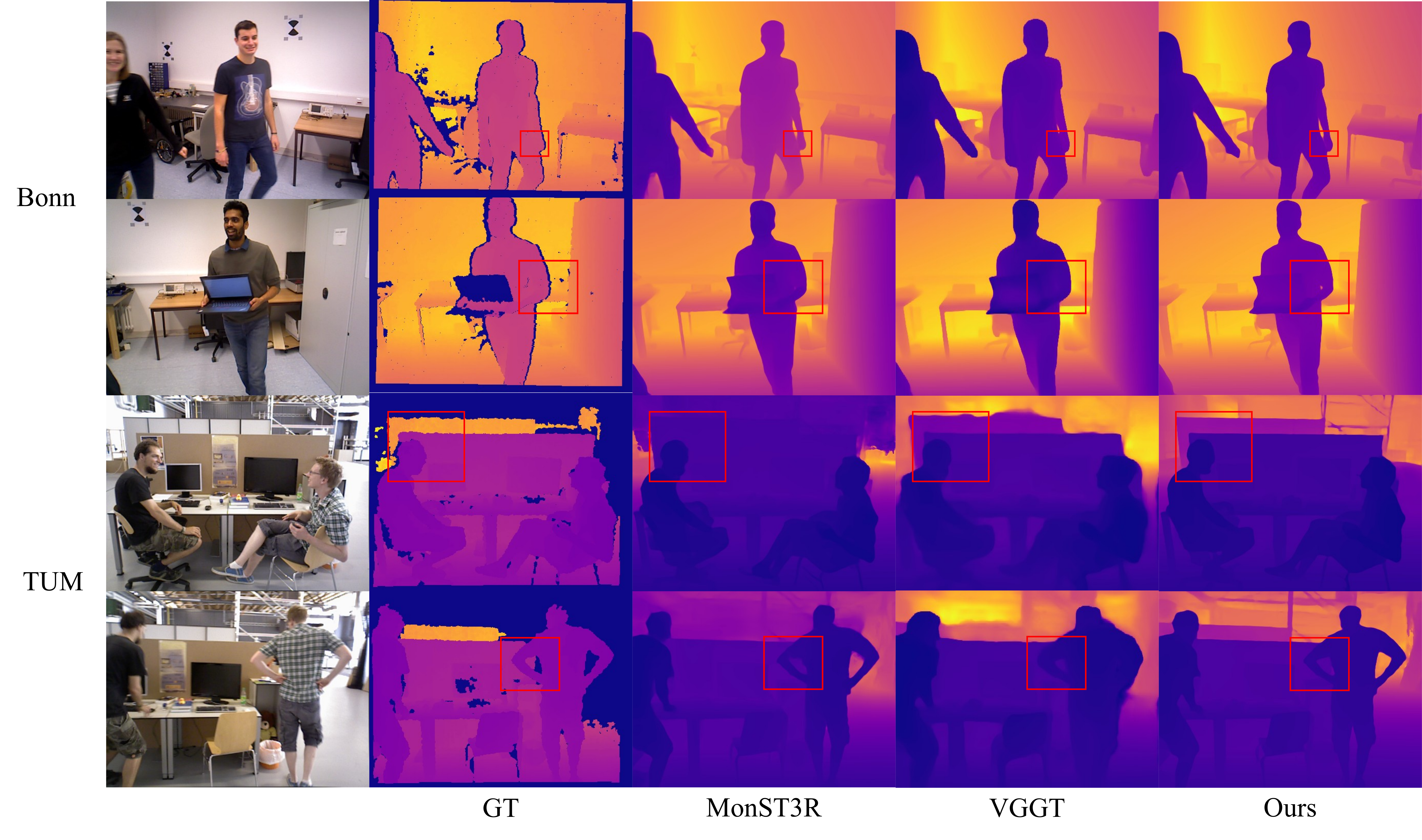}
    \caption{Qualitative Comparison on Bonn and Tum dynamic Datasets. The red boxes highlight the comparison regions, demonstrating that our method outperforms existing works.}
    \label{fig:Qualitative Comparison}
\end{figure*}
\subsection{Camera Pose Estimation}
\paragraph{Evaluating dataset.} For evaluation, we use three datasets, comprising two real-world datasets—Bonn, TUM Dynamics—and one synthetic dataset, GTA-IM.
\paragraph{Evaluation metrics.} We follow the evaluation protocol from previous works \cite{lu2024align3ralignedmonoculardepth} and report three metrics: Absolute Trajectory Error (ATE), Relative Translation Error (RTE), and Relative Rotation Error (RRE). ATE measures the absolute difference between estimated and ground truth camera trajectories. RTE and RRE measure the relative translation and rotation errors between consecutive frames, respectively.
\paragraph{Results.} Table~\ref{pose_estimation} shows the camera pose estimation results. Our HuPrior3R achieves the best performance on synthetic GTA-IM dataset across all metrics (ATE: 0.015, RTE: 0.010, RRE: 0.227), demonstrating significant improvements in temporal consistency and geometric coherence for human-centric scenes. On real-world datasets, our method shows competitive performance, achieving second-best ATE on TUM Dynamics (0.335) and second-best RTE on Bonn (0.285), while obtaining the best RRE on TUM Dynamics (0.430).

\subsection{Ablation Study}
\subsubsection{Prior Incorporation}
\vspace{-0.5em}
To validate the effectiveness of our Feature Fusion Module (FFM) in incorporating SMPL priors, we conduct a controlled ablation study on a representative sequence from the GTA dataset. We compare three model variants: our proposed FFM, a baseline without SMPL prior incorporation (only mono prior), and a naive fusion approach without FFM.

\textbf{Experimental Setup.} We train all models on 1,000 image pairs from the same GTA sequence for 50 epochs under identical training conditions. We evaluate three configurations:

\begin{itemize}
    \item \textbf{Only mono}: Uses only monocular depth features without any SMPL prior incorporation.
    \item \textbf{w/o FFM}: Incorporates SMPL priors through simple linear addition:
    \begin{equation}
    \hat{\mathbf{E}}_{\text{scene}}^{(l)} = \text{ZeroConv}(\hat{\mathbf{F}}_{\text{mono}}^{(l)})+ \text{ZeroConv}(\hat{\mathbf{F}}_{\text{smpl}}^{(l)}) +\mathbf{E}_{\text{base}}^{(l)}
    \end{equation}
    \item \textbf{w. FFM}: Our full model employing the multi-head attention-based gating mechanism described in Section~\ref{sec:method} to intelligently fuse multi-modal features.
\end{itemize}

\textbf{Results.} As shown in Table~\ref{table:ablation}, incorporating SMPL priors significantly improves depth estimation performance. Comparing "only mono" with our fusion approaches, we observe substantial improvements: Absolute Relative error drops from 0.054 to 0.034 (37\% improvement), and accuracy within the 1.25 threshold increases from 0.978 to 0.980. 

More importantly, our Feature Fusion Module demonstrates clear advantages over naive fusion. The FFM achieves better Absolute Relative error (0.034 vs 0.036) and higher accuracy (0.980 vs 0.978) compared to simple linear addition. While Log RMSE shows mixed results, with "only mono" achieving the best score (0.125), this metric can be sensitive to outliers and doesn't reflect the overall improvement in depth quality.

 \textbf{Qualitative Analysis.} Our qualitative evaluation demonstrates progressive improvements across different fusion strategies. The naive linear addition approach (w/o FFM) introduces artifacts in background regions where SMPL priors are irrelevant, leading to degraded depth estimation quality in non-human areas. This occurs because simple fusion cannot distinguish between regions where human body priors are beneficial versus regions where they introduce noise.

Unlike simple linear addition, our gating mechanism learns to selectively incorporate SMPL depth information when it provides reliable geometric constraints while downweighting it in regions where image-based features are more reliable. This adaptive fusion strategy prevents potential conflicts between different modalities and enables more robust depth estimation across the entire scene. Additional qualitative Figure~ref{} comparisons are provided in the supplementary material.

\begin{table}[!tbp]
\footnotesize
\setlength\tabcolsep{1.5pt}
  \caption{\textbf{Ablation study on Feature Fusion Module (FFM).}}%
  \centering
    \label{table:ablation}
  \begin{tabular}{c|ccc}
      \toprule
     Setting &Abs Rel$\downarrow$ & $\delta$$<$1.25$\uparrow$ & Log RMSE$\downarrow$\\
    \midrule
    only mono&0.042&0.978&\textbf{0.125}\\
    w/o FFM&\underline{0.037}&0.978&0.138 \\
    w. FFM&\textbf{0.035}&\textbf{0.980}&\underline{0.137} \\
      \bottomrule
  \end{tabular}
\end{table}
The results demonstrate that: (1) SMPL prior incorporation is essential for accurate human depth estimation; (2) naive fusion can introduce artifacts; and (3) our FFM provides the most effective integration strategy. The superior performance of FFM can be attributed to its ability to adaptively weight the contribution of human body priors based on scene context. Unlike simple linear addition, which treats all features equally, our gating mechanism learns to selectively incorporate SMPL depth information when it provides reliable geometric constraints while downweighting it in regions where image-based features are more reliable.

\begin{figure}[t]
    \centering
    \includegraphics[width=0.93\columnwidth]{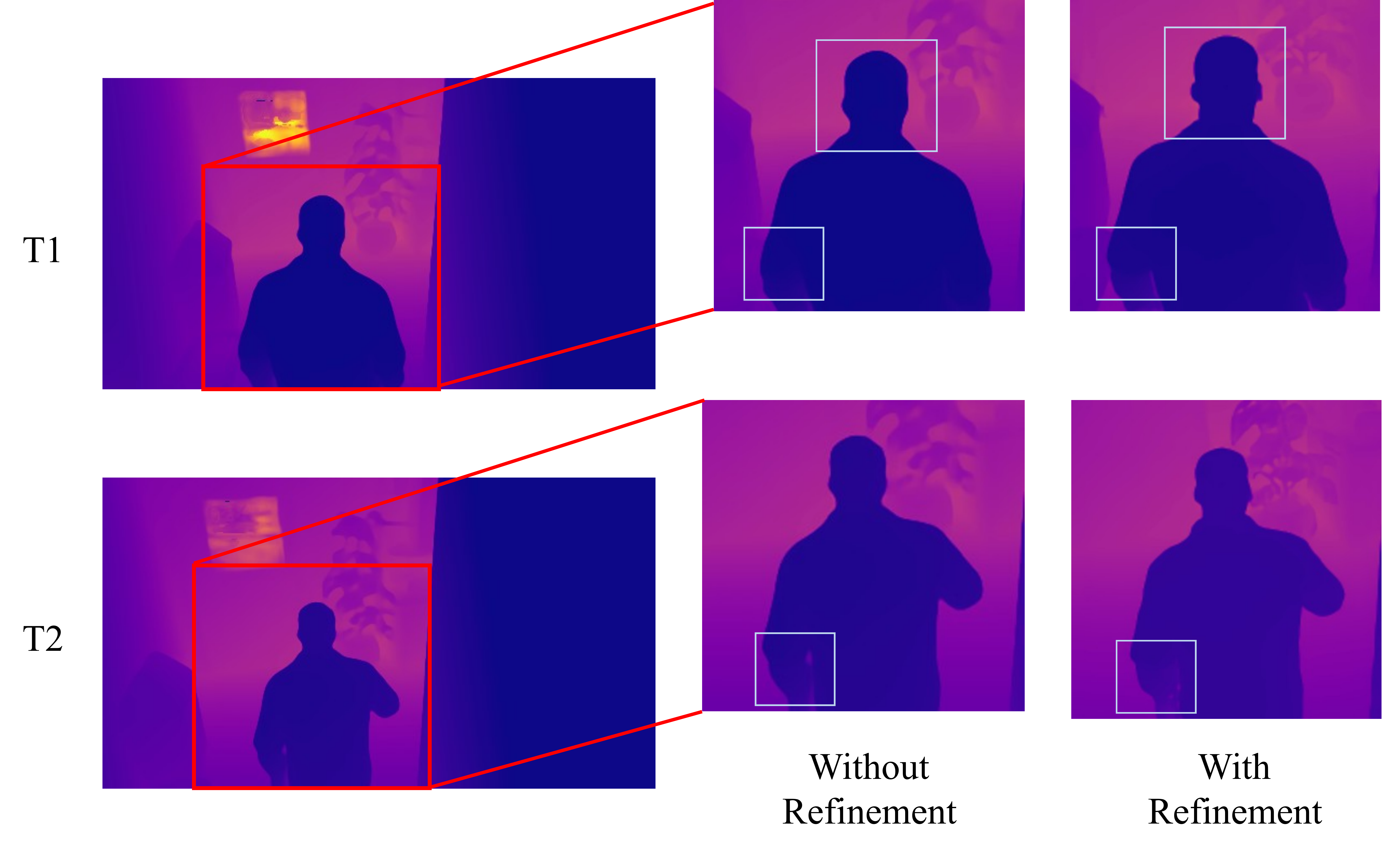} 
    \caption{Qualitative comparison of human reconstruction on HuPrior3R with and without refinement. Our refinement module produces more accurate human geometry and maintains better temporal consistency across frames in the cropped human regions.}
    \label{fig:Refinement}
\end{figure}
\vspace{-1em}
\subsubsection{Refinement}
\vspace{-0.5em}
To validate the effectiveness of our refinement strategy in improving human reconstruction quality and temporal stability, we conduct qualitative comparisons on human-centric regions. Figure 4 demonstrates the visual improvements achieved through our hierarchical refinement approach.
\vspace{-0.5em}
\paragraph{Qualitative Analysis.} As shown in Figure~\ref{fig:Refinement}, our refinement strategy yields significant improvements in human reconstruction quality. The refined results exhibit sharper anatomical details and more accurate geometric boundaries compared to the base model without refinement. More importantly, the refinement module enhances temporal consistency across video sequences, reducing flickering artifacts and maintaining stable human geometry throughout dynamic movements.

The improvements are particularly evident in challenging scenarios involving complex human poses and occlusions. Our hierarchical pipeline, which processes full-resolution images for overall scene geometry while applying strategic cropping and cross-attention fusion for human-specific detail enhancement, successfully preserves fine-grained human boundaries while maintaining global scene consistency. This demonstrates that our refinement approach effectively addresses the anatomical inconsistencies and boundary degradation issues commonly observed in existing methods.

%% file: sec/5_conclusion.tex
\section{Conclusion}
\label{sec:conclusion}

We proposed HuPrior3R, a novel approach for monocular dynamic video reconstruction in human-centric scenes that addresses anatomical artifacts and resolution degradation issues. Our method incorporates hybrid geometric priors combining SMPL human body models with monocular depth estimation for anatomically plausible reconstruction.

The hierarchical pipeline processes full-resolution images for scene geometry while applying strategic cropping and cross-attention fusion for human detail enhancement. Experiments on TUM Dynamics and GTA-IM datasets validate superior performance in resolving anatomical inconsistencies and boundary degradation.

%% file: sec/X_suppl.tex
\clearpage
\setcounter{page}{1}
\maketitlesupplementary
\section{More Feature Fusion Module Details}
\begin{figure}[h]
    \centering
    \includegraphics[width=1\columnwidth]{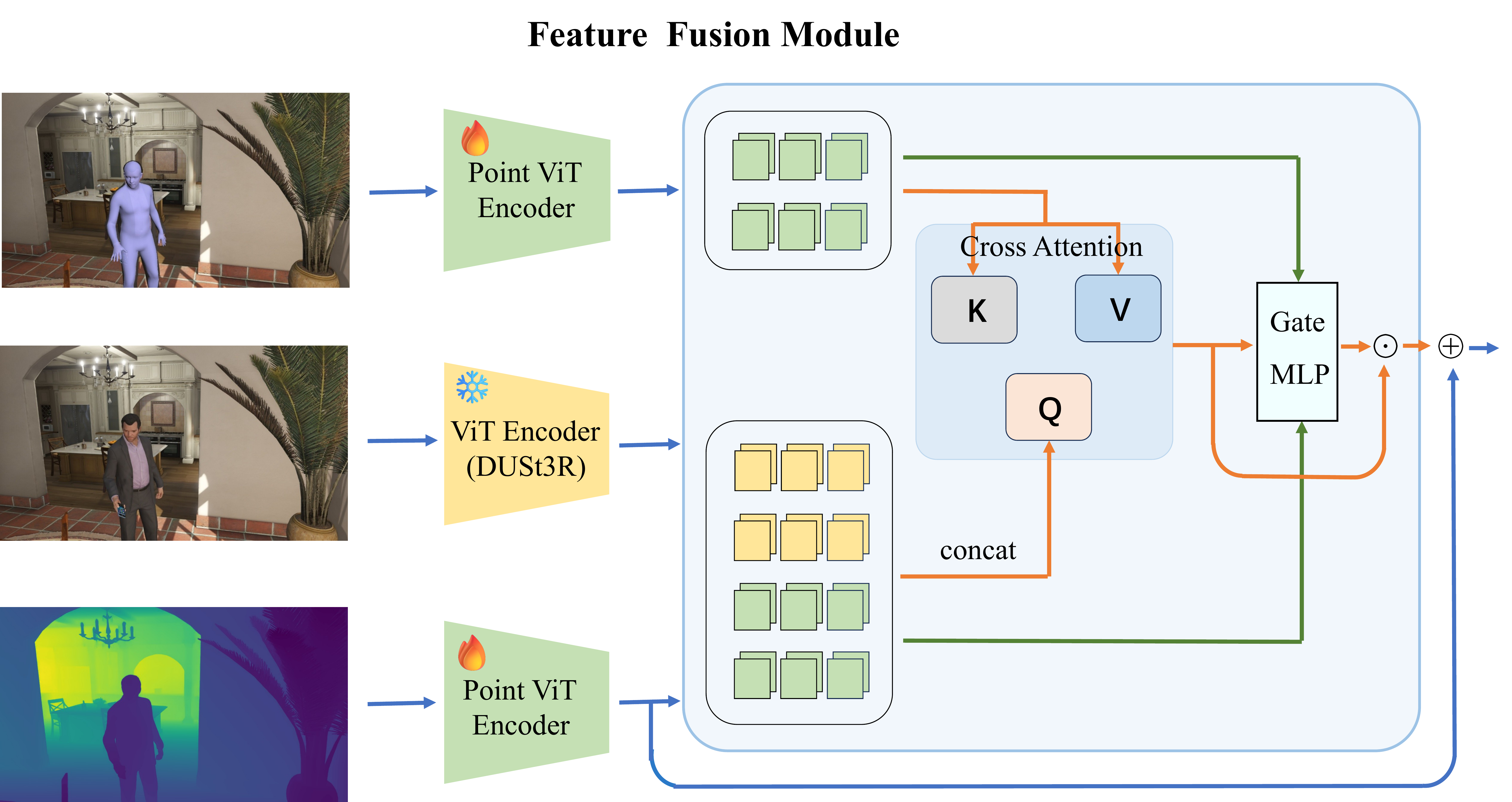} 
    \caption{Detail of Feature Fusion Module.}
    \label{fig:Feature Fusion Module}
\end{figure}

In our pipeline, We employ two separate feature fusion modules for image 1 and image 2, each structured as illustrated in Figure~\ref{fig:Feature Fusion Module}. The cross-attention mechanism within each module utilizes 8 attention heads.

\section{More Ablation Study Results}

\subsection{Qualitative Analysis of Prior Incorporation}

\begin{figure*}[t]
    \centering
    \includegraphics[width=1\linewidth]{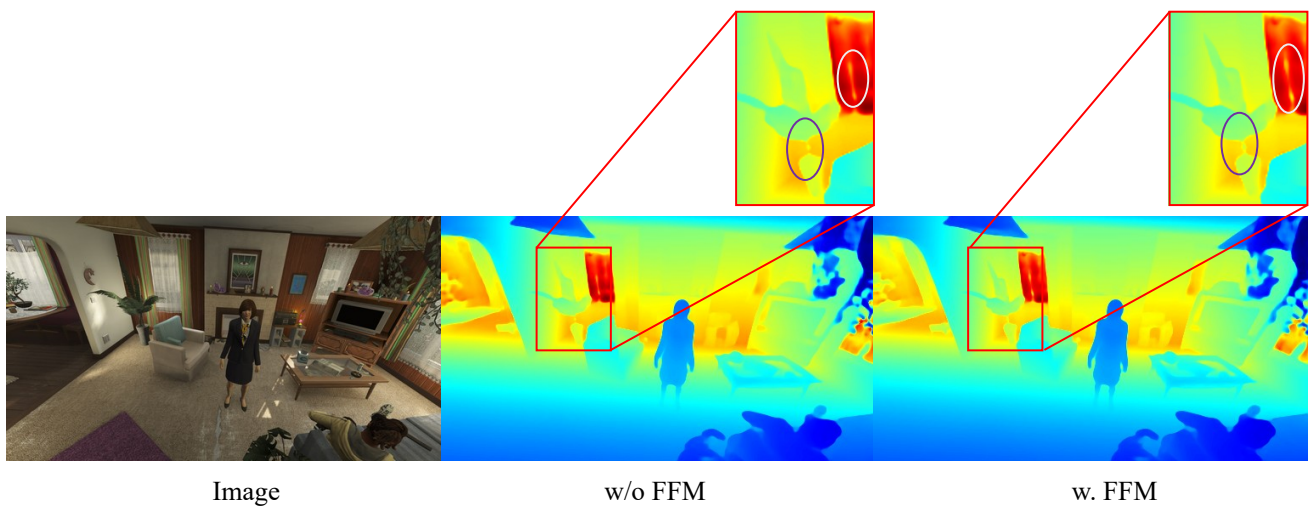} 
    \caption{Qualitative ablation study on Prior Incorporation. The naive fusion approach (w/o FFM) introduces artifacts in background regions where SMPL priors are irrelevant, while our FFM preserves background quality.}
    \label{fig:ablation_qualitative}
\end{figure*}

Figure~\ref{fig:ablation_qualitative} provides detailed qualitative comparisons of our Feature Fusion Module (FFM) against baseline approaches. The results demonstrate the progressive improvements achieved through our multi-modal fusion strategy.

\subsection{Depth Alignment Ablation Study}

To validate the necessity of aligning monocular and SMPL depth maps as described in Section 3.2.1, we conduct an ablation study comparing our RANSAC-based alignment approach with direct fusion of unaligned depths.

\textbf{Experimental Setup.} We compare two configurations on the same GTA sequence used in our FFM ablation study:
\begin{itemize}
    \item \textbf{w/o Alignment}: Directly incorporates raw monocular depth $\hat{\mathbf{D}}_{k}^{\mathrm{Mono}}$ and SMPL depth $\hat{\mathbf{D}}_{k}^{\mathrm{SMPL}}$ into the Feature Fusion Module without any scale or offset correction.
    \item \textbf{w. Alignment}: Our full approach using RANSAC-based linear transformation (Equations 1-2) to obtain aligned depth $\hat{\mathbf{D}}_{k}^{\mathrm{Aligned}}$ before feature extraction and fusion.
\end{itemize}

\textbf{Results and Analysis.} As shown in Table~\ref{table:ablation_align}, the absence of depth alignment severely degrades performance. Without proper alignment, significant scale and offset discrepancies between monocular and SMPL depth estimates create conflicting geometric signals during training. This leads to several critical issues:

\begin{itemize}
    \item \textbf{Prior rejection}: The network learns to treat SMPL depth as noise rather than meaningful geometric constraints, with the Feature Fusion Module suppressing rather than selectively incorporating SMPL features.
    \item \textbf{Cross-attention failure}: Scale inconsistencies prevent the multi-head attention mechanism from establishing meaningful correspondences between depth modalities.
    \item \textbf{Training instability}: Conflicting depth representations hinder effective multi-modal learning, resulting in substantially worse performance across all metrics.
\end{itemize}

The dramatic performance difference (Abs Rel: 5.12 vs 0.035) demonstrates that our RANSAC-based alignment is not merely a preprocessing convenience, but a fundamental requirement for effective multi-modal fusion. By ensuring consistent depth representation across estimation sources, the alignment step enables the network to properly utilize SMPL priors as complementary geometric information, validating the critical importance of this component in our pipeline.

\begin{table}[!tbp]
\footnotesize
\setlength\tabcolsep{1.5pt}
  \caption{\textbf{Ablation study on  Depth AlignmentM).}}%
  \centering
    \label{table:ablation_align}
  \begin{tabular}{c|ccc}
      \toprule
     Setting &Abs Rel$\downarrow$ & $\delta$$<$1.25$\uparrow$ & Log RMSE$\downarrow$\\
    \midrule
    w/o Align&5.12&0.084&1.834 \\
    w. Align&\textbf{0.035}&\textbf{0.980}&\textbf{0.137} \\
      \bottomrule
  \end{tabular}
\end{table}

\section{More Qualitative Results}
\subsection{Camera pose comparison}
\begin{figure*}[t]
    \centering
    \includegraphics[width=1\linewidth]{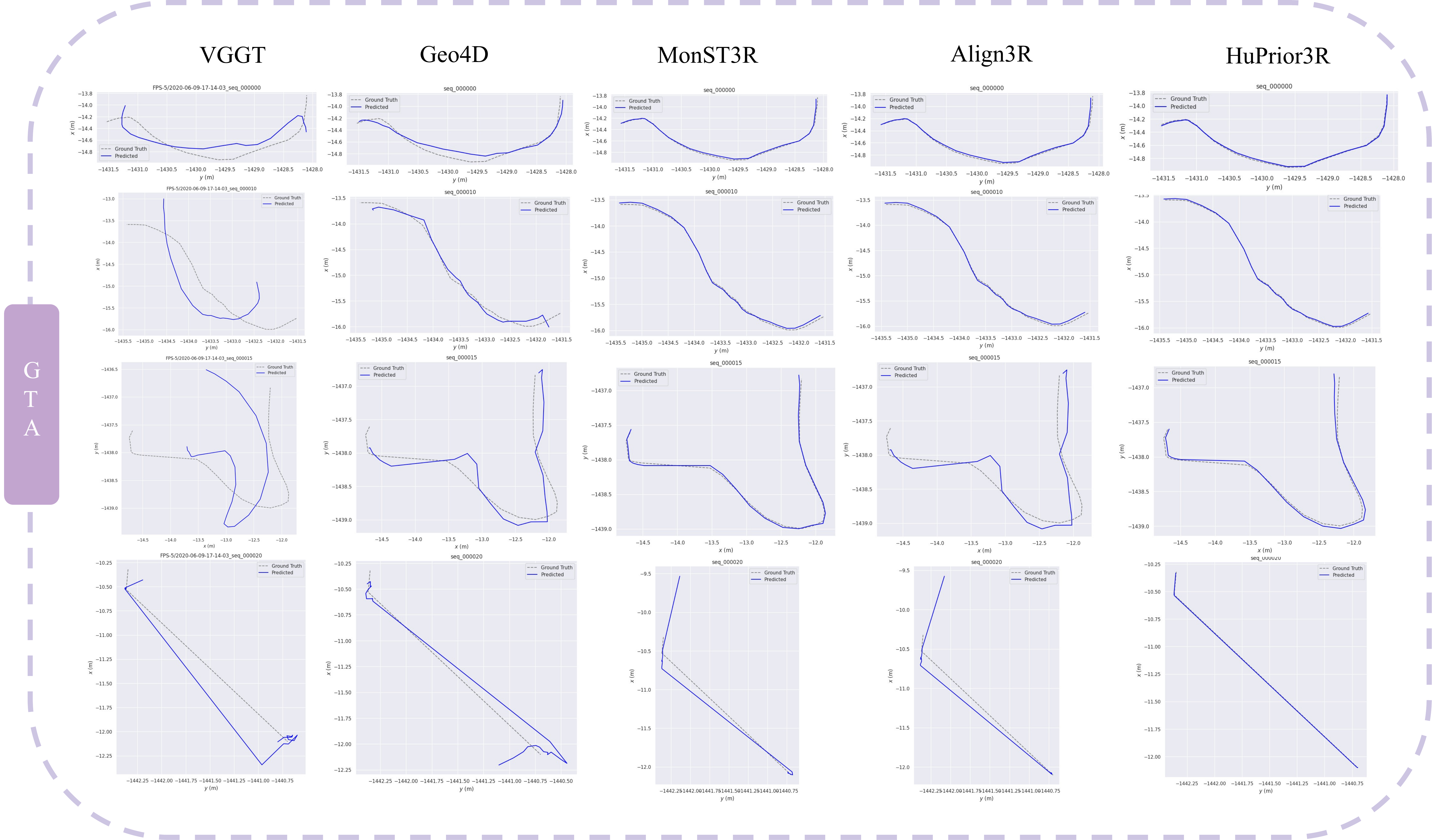} 
    \caption{Camera pose estimation comparison on the GTA-IM\cite{caoHMP2020} with existing work}
    \label{fig:pose_gta}
\end{figure*}

Figure~\ref{fig:pose_gta} demonstrates the camera pose estimation performance of our HuPrior3R method compared to existing approaches on the GTA-IM dataset. Our method achieves superior trajectory accuracy and temporal consistency, particularly in human-centric scenes.
\subsection{Dynamic point clouds}
\begin{figure*}[t]
    \centering
    \includegraphics[width=1\linewidth]{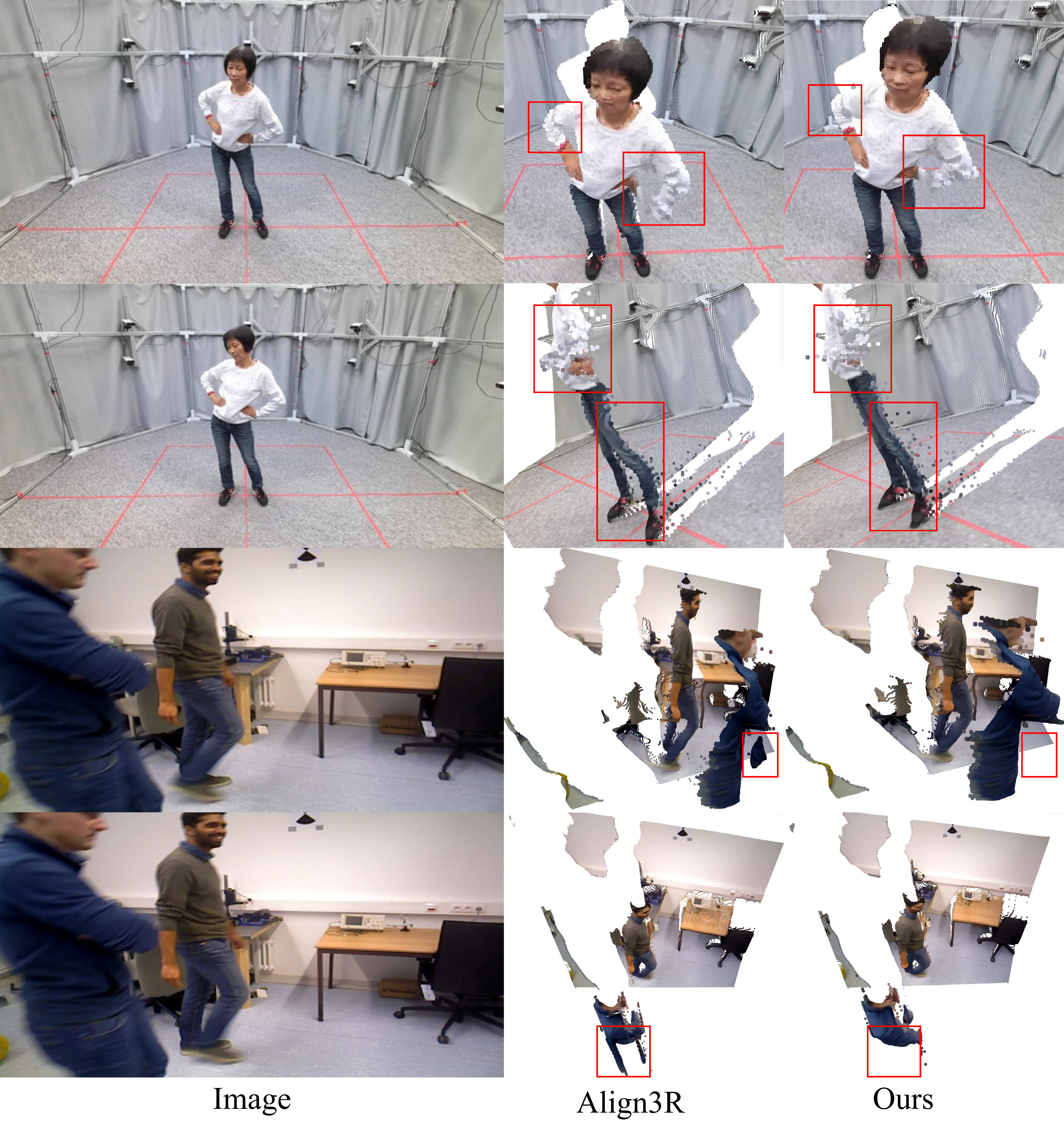} 
    \caption{Visualization of point clouds on the HuMMan \cite{cai2022humman} and Bonn \cite{10.1109/IROS40897.2019.8967590}. he visualizations highlight regions (marked in red boxes) where our approach produces more accurate and detailed human reconstructions compared to Align3R.}
    \label{fig:pointmap}
\end{figure*}

Figure~\ref{fig:pointmap} shows the 3D point cloud reconstructions on real-world datasets. Our approach produces more complete and geometrically consistent point clouds, especially in human body regions where SMPL priors provide valuable geometric constraints. The results demonstrate the effectiveness of our multi-modal fusion strategy in preserving both human body structure and background scene geometry.
